\documentclass[conference]{IEEEtran}
\usepackage{hyperref}
\hypersetup{
    colorlinks=true,
    linkcolor=orange,
    filecolor=magenta,      
    urlcolor=orange,
    pdftitle={Overleaf Example},
    pdfpagemode=FullScreen,
    }

\usepackage{cite}
\usepackage{amsmath,amssymb,amsfonts}
\usepackage{algorithmic}
\usepackage{graphicx}
\usepackage{textcomp}
\usepackage{xcolor}
\def\BibTeX{{\rm B\kern-.05em{\sc i\kern-.025em b}\kern-.08em
    T\kern-.1667em\lower.7ex\hbox{E}\kern-.125emX}}

\makeatletter
\def\section{\@startsection{section}{1}{\z@}{2.0ex plus 0.5ex minus 0.5ex}%
{0.7ex plus 0.5ex minus 0.5ex}{\normalfont\LARGE\bfseries\centering}}
\makeatother

\makeatletter
\def\subsection{\@startsection{subsection}{2}{\z@}{1.5ex plus 0.5ex minus 0.5ex}%
{0.5ex plus 0.5ex minus 0.5ex}{\normalfont\large\bfseries\centering}}
\makeatother

\begin{document}

\title{\textbf{Words of War:\\Exploring the Presidential Rhetorical\\Arsenal with Deep Learning}}

\author{
\IEEEauthorblockN{Wyatt Scott} 
\IEEEauthorblockA{\textit{School of Data Science} \\
\textit{University of Virginia}
}
\and
\IEEEauthorblockN{Brett Genz} 
\IEEEauthorblockA{\textit{School of Data Science} \\
\textit{University of Virginia}
}
\and
\IEEEauthorblockN{Sarah Elmasry} 
\IEEEauthorblockA{\textit{School of Data Science} \\
\textit{University of Virginia}
}
\and
\IEEEauthorblockN{Sodiq Adewole} 
\IEEEauthorblockA{\textit{School of Data Science} \\
\textit{University of Virginia}
}
}

\maketitle
\section*{\textbf{Abstract}}
In political discourse and geopolitical analysis, national leaders’ words hold profound significance, often serving as harbingers of pivotal historical moments. From impassioned rallying cries to calls for caution, presidential speeches preceding major conflicts encapsulate the multifaceted dynamics of decision-making at the apex of governance. This project aims to use deep learning techniques to decode the subtle nuances and underlying patterns of US presidential rhetoric that may signal US involvement in major wars. While accurate classification is desirable, we seek to take a step further and identify discriminative features between the two classes (i.e., interpretable learning).

Through an interdisciplinary fusion of machine learning and historical inquiry, we aspire to unearth insights into the predictive capacity of neural networks in discerning the preparatory rhetoric of US presidents preceding war. Indeed, as the venerable Prussian General and military theorist Carl von Clausewitz admonishes, “War is not merely an act of policy but a true political instrument, a continuation of political intercourse carried on with other means” (Clausewitz, 1832).

\section*{\textbf{Motivation}}
We aim to shed light on the interplay between the verbiage of national leaders and the inexorable currents of history that they set in motion. In addition to probing the efficacy of deep learning and natural language processing (NLP) while navigating the challenges inherent in the analysis of protracted textual corpora, we endeavor to examine how presidential rhetoric shapes, reflects and occasionally catalyzes the nation’s trajectory toward pivotal global events. We aim to gauge the impact of leaders’ orations on national decisions and international relations, furnishing novel insights and fresh perspectives on matters of global import.

Moreover, this interdisciplinary approach provides valuable tools for policymakers, historians, and the wider public. Deciphering the recurrent motifs within presidential addresses holds the potential to inform prognostication or influence forthcoming events, thereby exemplifying the enduring relevance of Clausewitzian principles in conjunction with contemporary technological innovations. In doing so, it bridges age-old theories with cutting-edge methodologies, fostering a more comprehensive understanding of how leaders adeptly frame their rhetoric to galvanize support for political endeavors. While impressive accuracy warrants attention and is important for a classification task as important as ours, we seek to make our model results interpretable; deep neural networks for classification are, to most, black boxes; we plan to use interpretable learning techniques to shed insight on how/why our models predict as they do.

\section*{\textbf{Methods}}
The data for this project comes from Kaggle, but the author scraped the data from The Miller Center at the University of Virginia (Lilleberg, 2020). We added a column to the dataset that represents our binary categorical response variable (War), indicating whether the US entered a major war within one year of the president’s speech. We encode an observation’s value for the War variable as 1 if the US entered a major war within one year of the president’s speech; otherwise, we encode the observation’s value for the War variable as 0. We derived wars’ start dates from the US Congressional Research Service (Torreon and Miller, 2024).

We perform some slight cleaning and preprocessing to set up the data for modeling. First, we checked for null values and found one missing transcript for a speech delivered by Thomas Jefferson on Nov. 8, 1808; we found the transcript via the Miller Center and added it to the dataset. Next, because the first war we consider (First Barbary War) started in 1801, we filter the dataset to speeches dated after 1800.

Several transcripts end with the president’s signature; we remove the signature text from the transcripts column given that the president is identifiable from the president column and that text is not important for our modeling purposes. The transcripts also contain instances of long integers and floating point numbers when a president describes various treasury and debt statistics, for example. We remove floating point numbers and integers from the transcripts. Additionally, we convert the transcripts to lowercase and remove punctuation.

After cleaning the data and adding our response variable, the dataset contains 964 observations and exhibits significant class imbalance. There are 883 observations classified as War = 0 and 81 classified as War = 1; roughly 92\% of the speeches were not delivered within one year of the US entering a major war. We use the Synthetic Minority Over-sampling Technique (SMOTE) to balance the classes, and, as the authors suggest, we combine SMOTE with random undersampling of the majority class (Chawla et al., 2002). We combine these transformations into a single pipeline.

With the classes relatively balanced and the text minimally cleaned, we now convert the text data into a format suitable for our modeling purposes.

We leverage a pre-trained Bidirectional Encoder Representations from Transformers (BERT) model to tokenize and vectorize the raw text data, converting the speeches into fixed-length vectors that we pass as inputs to our models (Devlin et al., 2019). We experiment with various model architectures using a binary cross-entropy loss function, and we evaluate model performance across accuracy, F1-Score, and Area Under the Receiver Operating Characteristic Curve (AUC-ROC). The models train on 80\% of the data; we use half of the remaining 20\% for validation and half for testing. We trained each model for ten epochs using batches of size 32.

We battled with shape mismatches when trying to feed the vectorized representations into the BERT model because we stacked the predictor features before applying the resampling pipeline, so we set up a separate pipeline to transform the text data for BERT. In this second pipeline, we use the same approach as before, but we append the input IDs and attention masks to lists so they can be directly accessed during training and evaluation.

The models we experiment with include:
\vspace{1pt}
\begin{enumerate}
    \item Multilayer Perceptron (MLP)
    \item Recurrent Neural Network (RNN) with Long Short-Term Memory (LSTM)
    \item LSTM with Attention
    \item BERT
\end{enumerate}
\vspace{1pt}

\section*{\textbf{Experiments and Results}}

This section describes our models, interpretable learning approaches, and results.

\subsection*{\textbf{Model Architectures and Performance}}

\textbf{MLP:} Our MLP consists of two dense hidden layers with ReLU activation followed by dropout regularization and an output layer with a sigmoid activation function. We apply L2 regularization of 0.01 to the kernel weights in all dense layers to prevent overfitting. When compiling the model, we use the Stochastic Gradient Descent optimizer with a learning rate of 0.001 and Nesterov momentum of 0.99.

\vfill\break

\begin{figure}[ht]
    \centering
    \includegraphics[width=1\linewidth]{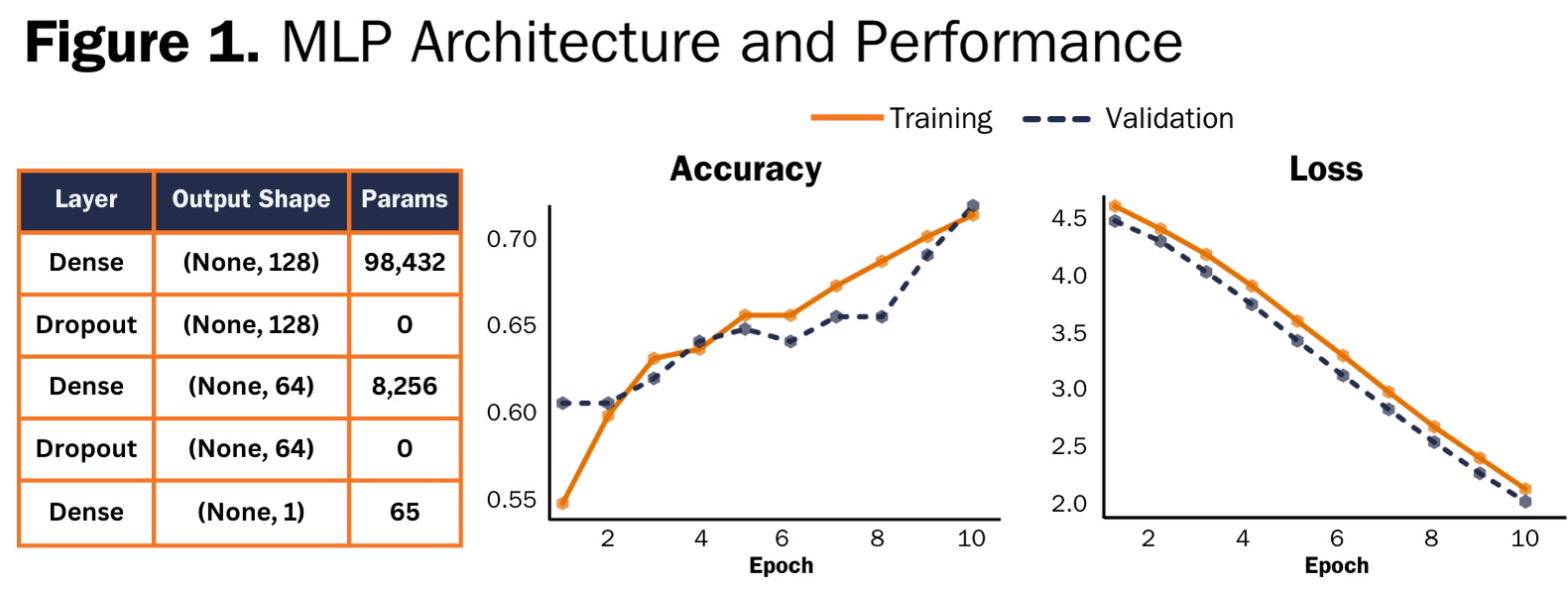}
\end{figure}

The MLP performs relatively well; the training and validation accuracy steadily improve, for the most part, and surpass 0.7 by epoch ten, and the training and validation loss steadily decrease.

\textbf{RNN with LSTM:} In our second model, we reshape the input data to include a timestep dimension before it’s fed into the LSTM layer, allowing the model to effectively capture temporal dependencies in the input data. With 128 units, the LSTM layer utilizes hyperbolic tangent activation, Glorot uniform, and orthogonal initializers, along with dropout of 0.1 and recurrent dropout of 0.1 for regularization. Next comes a densely connected layer consisting of 64 units with ReLU activation, He normal initialization, and L2 regularization of 0.1. We added a dropout layer to apply further regularization and mitigate overfitting. Given that we’re performing binary classification, the final layer is a dense output layer with a sigmoid activation function. We apply L2 regularization to the kernel weights in both dense layers to further prevent overfitting. When compiling the model, we use the Adam optimizer with a learning rate of 0.001.

\begin{figure}[ht]
    \centering
    \includegraphics[width=1\linewidth]{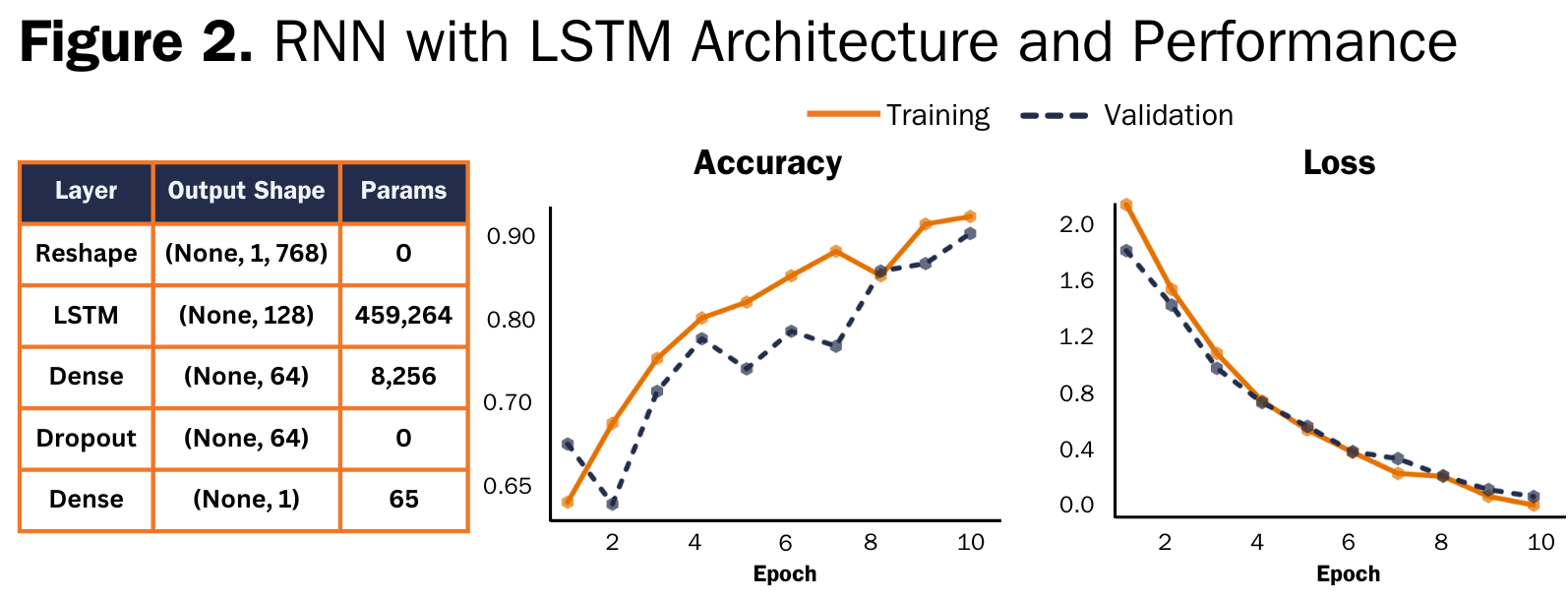}
\end{figure}

The RNN architecture with an LSTM layer performs better than the MLP; although the training and validation accuracy fluctuate somewhat, they steadily increase and reach over 0.9 by epoch ten. The training and validation loss steadily decrease across epochs.

\textbf{LSTM with Attention:} This model architecture is the same as the previous model except that it includes a custom attention layer between the LSTM layer and the first dense layer that dynamically weighs the input sequence elements based on their importance. As with the second model, we use the Adam optimizer with a learning rate of 0.001.

\begin{figure}[ht]
    \centering
    \includegraphics[width=1\linewidth]{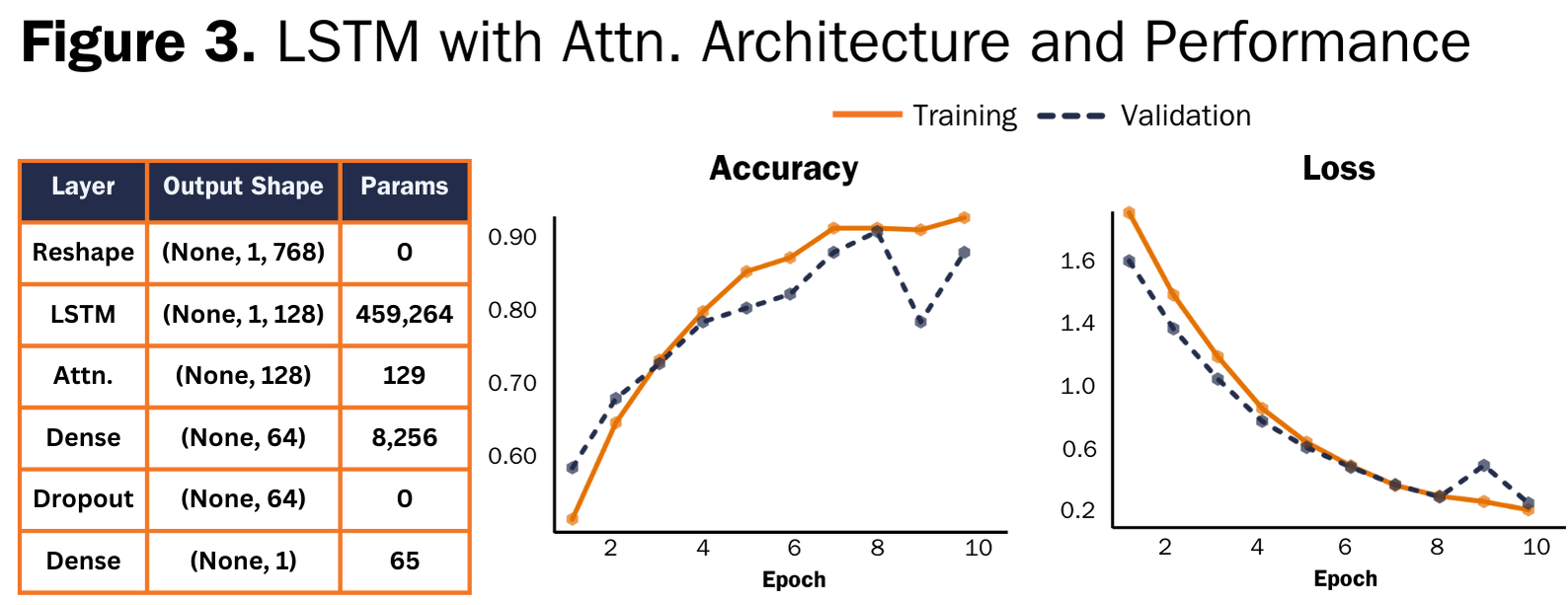}
\end{figure}

\vfill\break
Adding the attention layer seems to have improved performance compared to the previous two models. We observe the training and validation accuracy increasing steadily, except for a drop in validation accuracy in epoch nine. The training and validation loss decrease steadily and barely diverge.

\textbf{BERT:} The fourth model, fine-tuned on our dataset, utilizes self-attention mechanisms to process and analyze text segments in relation to their broader context within each speech. In training this model, we switch from the binary cross-entropy loss function to sparse categorical cross-entropy and use the Adam optimizer, specifying several hyperparameter values: 

\begin{itemize}
    \item learning rate = 3e-7
    \item $\beta_1$ = 0.9
    \item $\beta_2$ = 0.999
    \item $\epsilon$ = 1e-08
    \item clipnorm = 2.0
\end{itemize}

\begin{figure}[ht]
    \centering
    \includegraphics[width=1\linewidth]{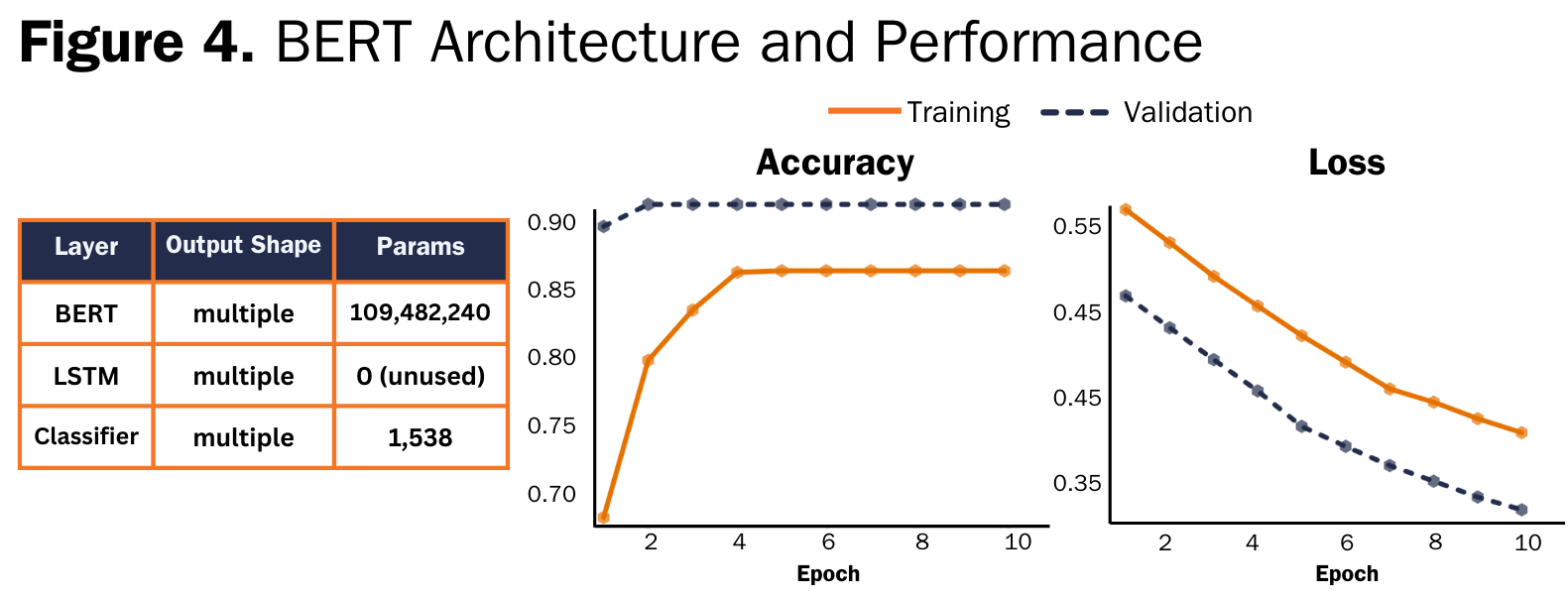}
\end{figure}

The fine-tuned BERT model effectively converges to high accuracy levels, showcasing its scalability despite the computational demands. The training and validation accuracy increase over the first few epochs but remain nearly constant thereafter, while the training and validation loss drop consistently.

The RNNs and BERT perform best on the training and validation sets, at least in terms of training and validation accuracy. BERT quickly reaches training and validation accuracy of over 0.9, while the RNNs take longer to get there. However, BERT takes much longer to train. Next, we use AUC-ROC and F1-Scores to compare model performance on the test dataset.

\vfill\break

\begin{figure}[ht]
    \centering
    \includegraphics[width=1\linewidth]{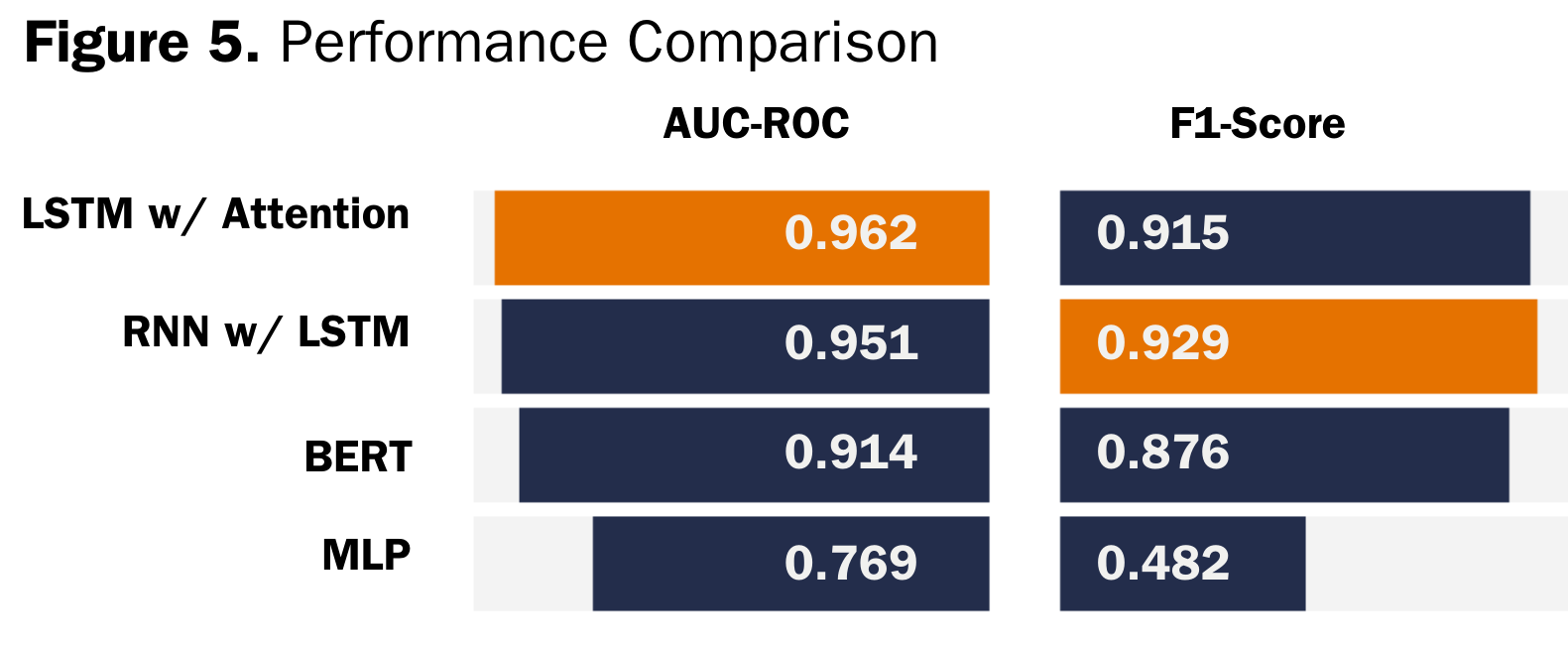}
\end{figure}

The RNNs and BERT all achieve AUC-ROC values over 0.9, although the LSTM with Attention has the highest value at 0.982, and the RNN with LSTM slightly outperforms the others in terms of F1-Score, achieving 0.929.

The MLP achieves a comparatively decent AUC-ROC but a much lower F1-Score, indicating that the classifier performs poorly when attaching heightened importance to false positives and false negatives.

\subsection*{\textbf{Interpretable Learning}}

While accurate classification is desirable, we also try to identify discriminative features between the two classes. We use three approaches to help interpret our models and results.

In our first approach, we extract and analyze the attention weights from our third model. To do so, we create another model using the Model class, specifying the same inputs as our third model but setting the output to that of the attention layer. This allows us to extract the attention weights, providing insights into how the attention mechanism weighs different parts of the input sequence.

\begin{figure}[ht]
    \centering
    \includegraphics[width=1\linewidth]{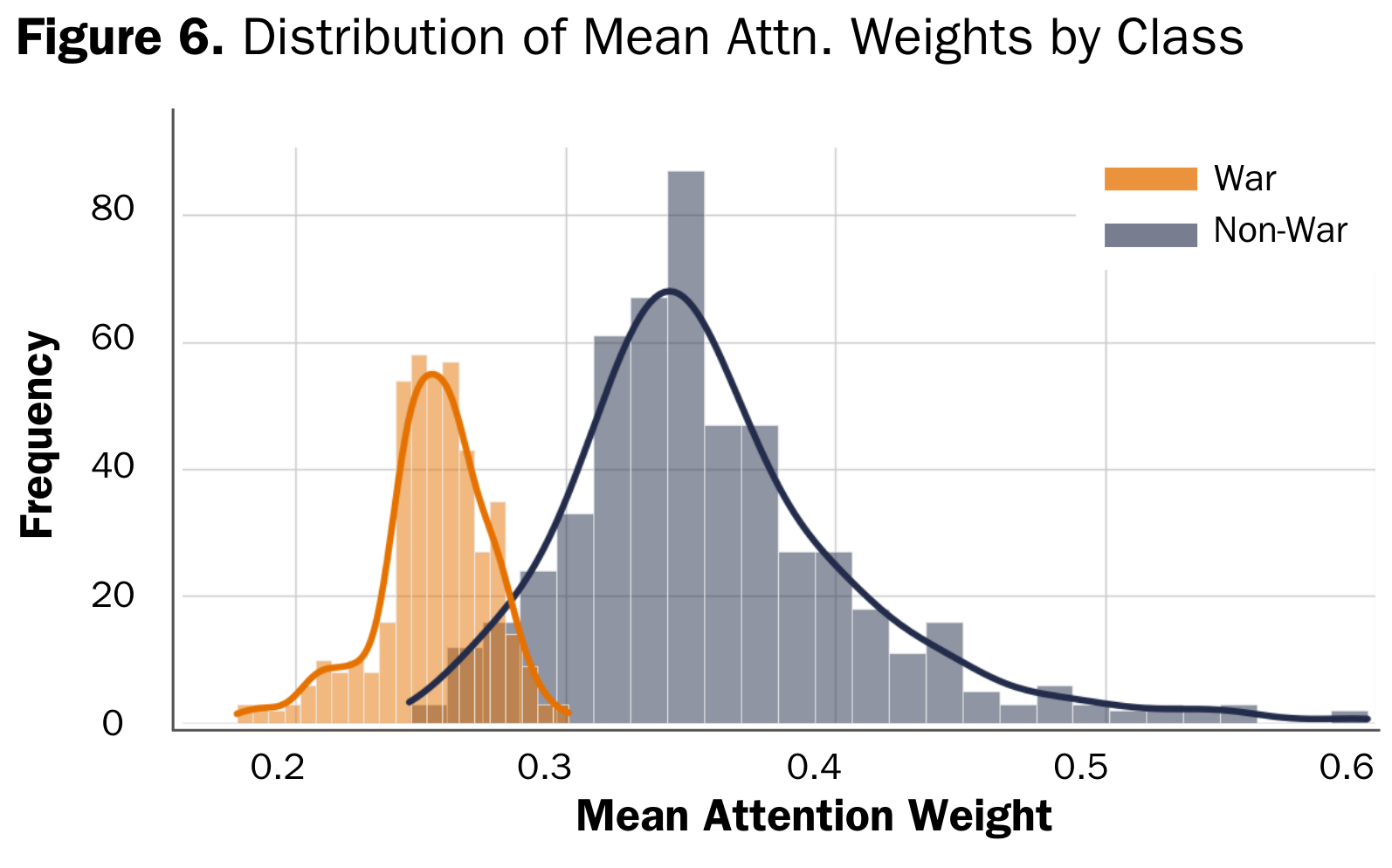}
\end{figure}

The x-axis of the distribution plot above represents mean attention weights, which indicate the average importance that the attention mechanism assigned to different pieces of the input sequences. The y-axis represents the frequency of sequences with a particular mean attention weight. The plot allows us to compare the mean attention weight distributions between the two classes; we observe some overlap but reasonably clear separation between the distributions of mean attention weights for the two classes, suggesting that the attention mechanism effectively captures differences between the classes.

For our second approach, we use the Local Interpretable Modeling-agnostic Explanations (LIME) package to help explain predictions from our second model (Ribeiro et al., 2016). Approximating our complex model via a local linear explanation model enables us to analyze and visualize the influence of individual features on prediction outcomes, helping identify key attributes that distinguish between classes and providing a basis for deeper analysis and justification of the model’s decisions.

\begin{figure}[ht]
    \centering
    \includegraphics[width=1\linewidth]{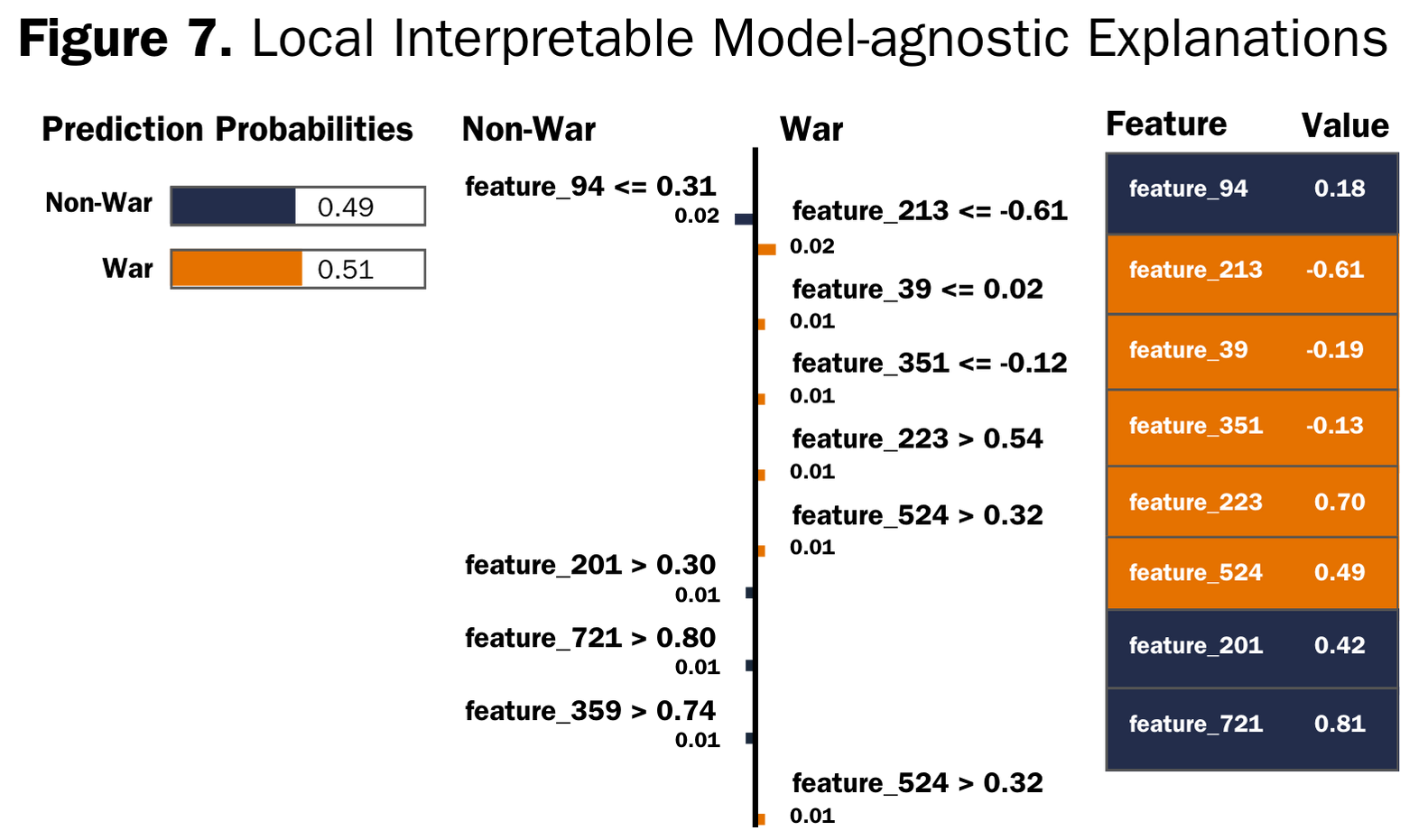}
\end{figure}

The chart above shows the dimensions that contributed most to a single prediction from the model; the bars indicate magnitude and whether the feature influenced the model toward or away from a prediction of War = 1. Investigating local explanations can provide insight into whether or not the model’s decisions align with human decision-making.

The third way we add interpretability is by employing the SHapley Additive exPlanations (SHAP) package to visualize feature importance values from the second model (Lundberg and Lee, 2017). In contrast with LIME, SHAP values explain how features affect a model globally.

\begin{figure}[ht]
    \centering
    \includegraphics[width=1\linewidth]{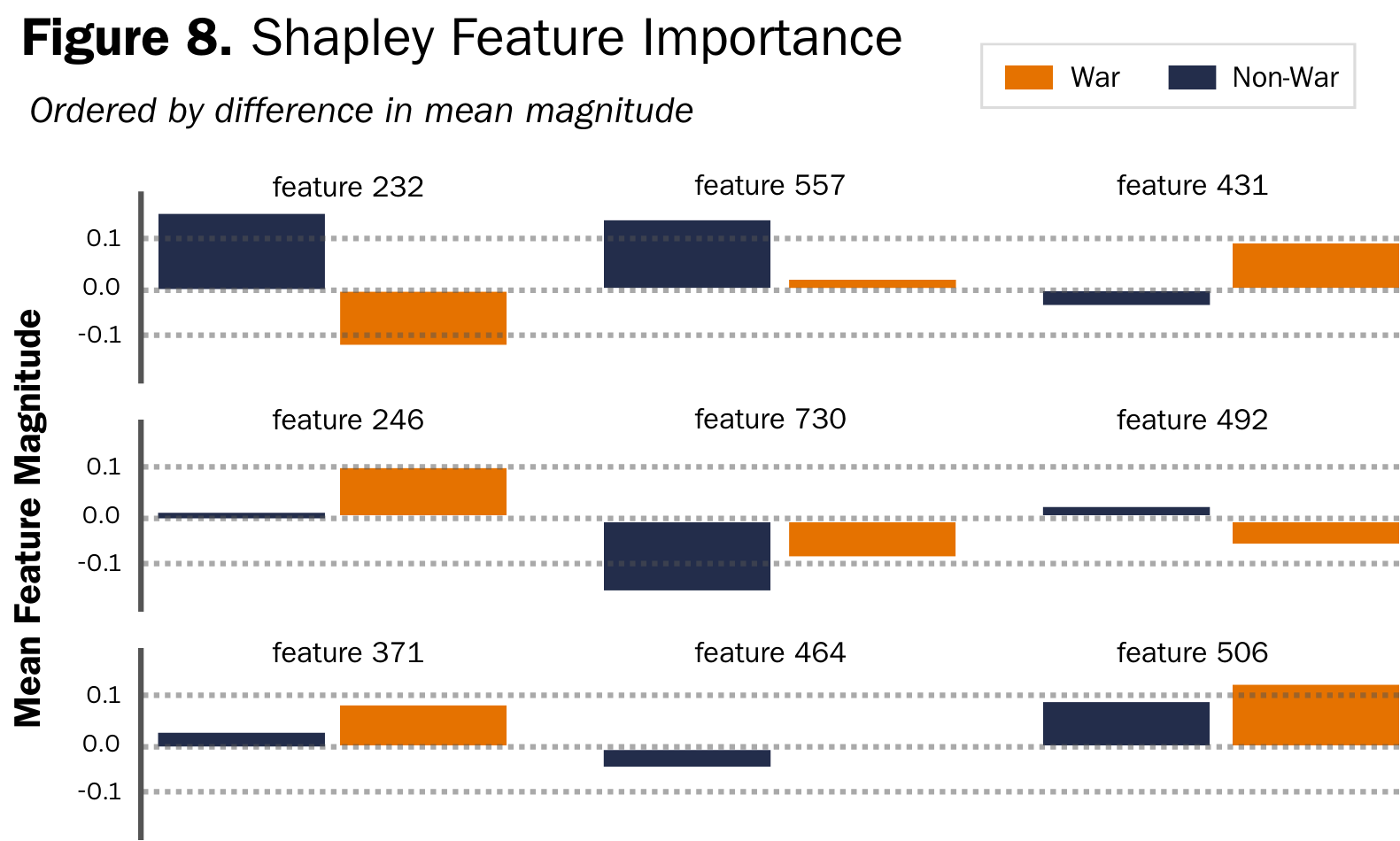}
\end{figure}

The visualization illustrates the most influential features SHAP identified for our second model, ranked by the largest mean magnitude associated with war predictions. By comparing SHAP with LIME, we observe that the key features influencing local predictions often differ significantly from those impacting global outcomes. This contrast highlights the unique insights each method brings to model interpretability.

\section*{\textbf{Conclusions}}

Our experiments not only demonstrate the potential for deep learning techniques to reveal patterns in US presidential rhetoric but also hint at their predictive power in determining involvement in future wars. The diverse neural network architectures we constructed and the pre-trained BERT model we utilized show that gated RNNs and transformer-based architectures can accurately classify text inputs of varying lengths, even in the face of extensive raw texts.

\subsection*{\textbf{Future Research}}
Exciting avenues for future research in this area could include experimentation with more advanced transformer models for classification as well as different language encoding techniques, such as sub-word tokenization. These explorations hold the promise of further enhancing our understanding and application of deep learning in text analysis.

Given that the texts of presidential speeches are longer, future research can experiment with emerging techniques to overcome the input sequence length limitation of powerful transformer-based models like BERT. BERT’s self-attention mechanism, for example, can process a maximum of 512 tokens. Overcoming such limitations requires careful preprocessing; for instance, researchers have explored employing truncation, chunking, etc. (Dong et al., 2022; Park et al., 2022). Other newer approaches, like BigBird and Longformer, use sparse attention mechanisms with larger maximum token limits, and others explore fine-tuning BERT to work with longer text data, including ChunkBERT (Jaiswal and Milios, 2023) and BERT For Longer Texts (BELT) (Brzozowski, 2023). Future research on our topic of focus would benefit from experimenting with similar approaches and evaluating model performance when the inputs capture most, if not all, of the longer-form texts.

Research has shown that the structure of the BERT-based gated approaches, which use a fully connected encoding unit and apply the gate mechanism to update state memory, are computationally inefficient given the quadratic time complexity of applying self-attention in long-text modeling. A recent paper (Li et al., 2023) proposes addressing these issues using what the authors refer to as a Recurrent Attention Network (RAN). The RAN model uses positional multi-head self-attention on local windows for dependency extraction and employs a Global Perception Cell (GPC) vector to propagate information across windows, concatenated with tokens in subsequent windows. The GPC vector acts as a window-level contextual representation and maintains long-distance memory, enhancing local and global understanding. Additionally, a memory review mechanism allows the GPC vector from the last window to serve as a document-level representation for classification tasks. Thus, future research on our topic of interest might look to leverage similarly powerful transformer-based models while optimizing efficiency.

There is much room for improving interpretability in the field of deep learning generally but more specifically in the context of these larger models. Researchers recently developed an approach called ProtoryNet, which makes predictions by finding the most similar prototype for each sentence in a sequence and feeding an RNN backbone with the proximity of each sentence of the corresponding active prototype. The RNN backbone then captures the temporal pattern of the prototypes, which the authors refer to as ‘prototype trajectories.’ These trajectories enable intuitive, find-grained interpretation of the RNN model’s reasoning process (Hong et al., 2023). Future research in long-text modeling (among other topics) might try to leverage ProtoryNet and other emergent approaches to increase model explainability and shine some light on the ‘black box’ of these models’ decision-making.

If we were to spend more time and expand our analysis, we would leverage the Party variable to examine whether partisan differences exist in war rhetoric or if any differentiable patterns exist when including party affiliation as an input feature.

Another way we might augment our research is by including speeches from the leaders of many countries operating under differing government structures with varying degrees of openness. The context would change slightly in that, for the current study, we operate under the assumption that US presidents need to get buy-in from citizens and massage the national psyche to support their cause; otherwise, the president risks losing power. The same assumption would likely not hold, or at least would need to be adjusted, for authoritarian regimes. Expanding the dataset to include speeches from different countries with different forms of government may open interesting avenues for future research into national leaders’ rhetoric and accountability.


\end{document}